\documentclass[10pt,twocolumn,letterpaper,runningheads]{article}

\usepackage[pagenumbers]{mypaper}      %

\usepackage{svg}
\usepackage{multirow}
\usepackage{makecell}
\usepackage{diagbox}
\usepackage{wrapfig}
\usepackage{pgfplots}
\usepackage{pifont}
\newcommand{\cmark}{\ding{51}}%
\newcommand{\xmark}{\ding{55}}
\usepackage[switch]{lineno}

\usepackage{placeins}

\newcommand{\affspace}{\hspace{2em}}

\definecolor{mypaperblue}{rgb}{0.21,0.49,0.74}
\usepackage[pagebackref,breaklinks,colorlinks,allcolors=mypaperblue]{hyperref}
\pgfplotsset{width=10cm,compat=1.9}

\title{Learn Your Scales: Towards Scale-Consistent Generative Novel View Synthesis}

\author{
Fereshteh Forghani\textsuperscript{1}
\and
Jason J.~Yu\textsuperscript{1}
\and
Tristan Aumentado-Armstrong\textsuperscript{1,3}
\and
Konstantinos G.~Derpanis\textsuperscript{1,2,3}
\hspace{+1.5em}
\vspace{+10pt}Marcus A.~Brubaker\textsuperscript{1,2,4} \\
{\normalsize \textsuperscript{1}York University \affspace
\textsuperscript{2}Vector Institute for AI \affspace
\normalsize\textsuperscript{3}Samsung AI Centre Toronto \affspace
\textsuperscript{4}Google DeepMind}\\
{\tt\small \{forghani,jjyu,marcus.brubaker,kosta\}@yorku.ca, tristan.a@samsung.com}\\
}

\begin{document}
\maketitle

\begin{abstract}
Conventional depth-free multi-view datasets are captured using a moving monocular camera without metric calibration.
The scales of camera positions in this monocular setting are ambiguous.
Previous methods have acknowledged scale ambiguity in multi-view data via various ad-hoc normalization pre-processing steps, but have not directly analyzed the effect of incorrect scene scales on their application.
In this paper, we seek to understand and address the effect of scale ambiguity when used to train generative novel view synthesis methods (GNVS).
In GNVS, new views of a scene or object can be minimally synthesized given a single image and are, thus, unconstrained, necessitating the use of generative methods.
The generative nature of these models captures all aspects of uncertainty, including any uncertainty of scene scales, which act as nuisance variables for the task.
We study the effect of scene scale ambiguity in GNVS when sampled from a single image by isolating its effect on the resulting models and, based on these intuitions, define new metrics that measure the scale inconsistency of generated views.
We then propose a framework to estimate scene scales jointly with the GNVS model in an end-to-end fashion.
Empirically, we show that our method reduces the scale inconsistency of generated views without the complexity or downsides of previous scale normalization methods.
Further, we show that removing this ambiguity improves generated image quality of the resulting GNVS model.
\end{abstract}
    
\vspace{-10pt}
\section{Introduction}
\label{sec:intro}

\begin{figure}[t]
  \centering
   \includegraphics[width=\linewidth]{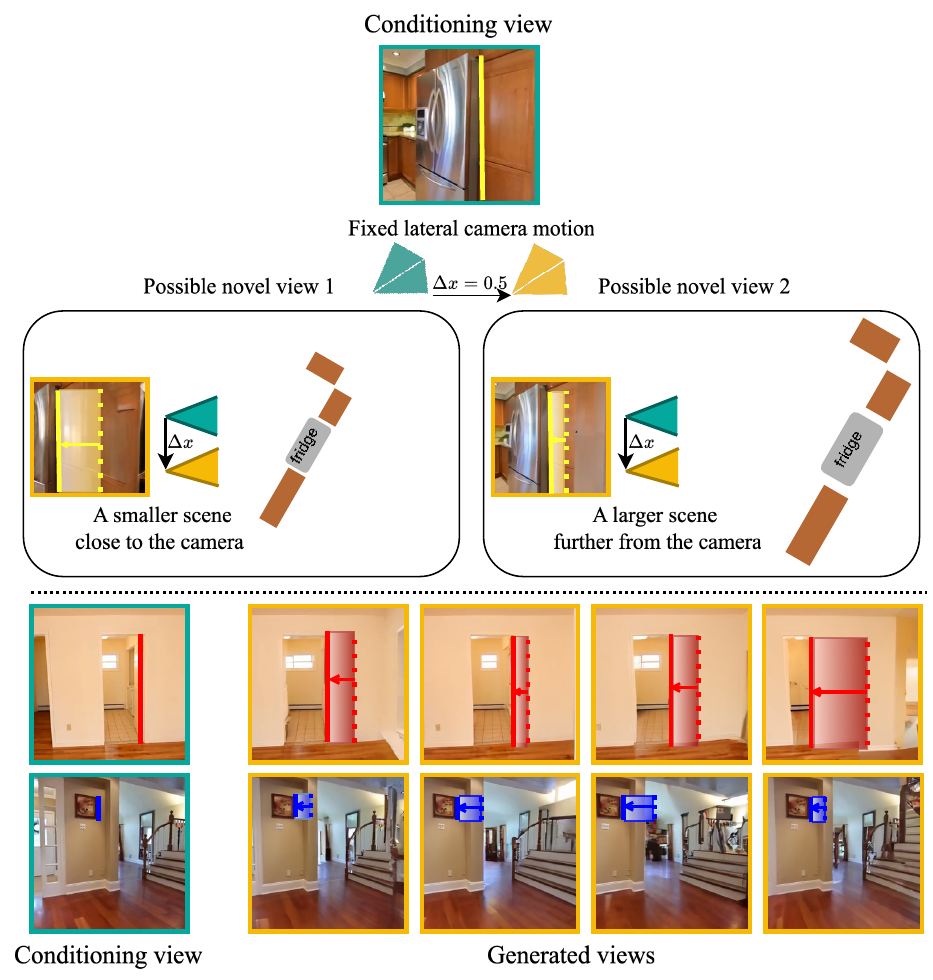}
   \caption{
   \textbf{Scale ambiguity and inconsistency in GNVS.}
   (Top) Two novel views are independently sampled using the same conditioning and camera motion, $\Delta x$.
   Samples exhibit different disparities due to uncertainty over scene scale.
   Here, we depict the plausible top down scene layouts of these samples in the boxes.
   This uncertainty occurs when GNVS models are trained with inconsistently calibrated data.
   (Bottom) Additional samples and scenes in the same setting are shown where a salient edge is highlighted to show the different disparities in the generated views.
   }
   \label{fig:SA_VO}
\end{figure}

A central task of visual perception is interpreting the 3D structure of 2D images, as geometric information is lost in the perspective projection process.
Multiview vision, such as uncalibrated binocular stereo, can recover substantial 3D structure \cite{hartley2003multiple}, but cannot obtain \textit{metric} geometry, 
due to the ``scale ambiguity'' inherent to images (see Fig.~\ref{fig:SA_VO}). 
This ambiguity complicates many 3D computer vision tasks, such as novel view synthesis (NVS), where uncalibrated data introduces variability in the perceived sizes and distances of objects.
Metric monocular depth estimation implicitly leverages visual cues to estimate scene scale but remains highly challenging 
(e.g., \cite{piccinelli2024unidepth,guizilini2023towards,bochkovskii2024depth}).
Similarly, for novel view synthesis (NVS) datasets (e.g., \cite{realestate10k}), precise metric calibration is generally not available, resulting in inconsistent scales across multiview scenes.
This scale uncertainty is a nuisance factor that can limit scene interpretation and hurt downstream tasks like NVS.
Here, we study the impact of metric scale uncertainty and examine how to mitigate it.

In particular, we consider the generative NVS (GNVS) task conditioned on a single view.
Standard ``interpolative'' NVS settings (e.g., NeRFs \cite{mildenhall2020nerf} and 3DGS \cite{kerbl20233d}) operate with many images, resolving scene geometry (up to a global scale) and allowing NVS for that scene in its arbitrary coordinate system.
In contrast, a GNVS model faces an under-determined scenario: given a single observed image and a trajectory of camera parameters, it must generate novel views for that camera path. 
Inconsistent scales in the data necessarily create uncertainty since the model can no longer assume a fixed size for objects -- that is, how can the model know what the camera will see if it does not know the scale of the scene relative to the given camera parameters?
This is readily observable in existing models (see Fig.~\ref{fig:SA_VO}), as the increased entropy of generated novel views.
We illustrate this phenomenon via the shifting of the fridge in Fig.~\ref{fig:SA_VO} (top), where two samples of a novel view are generated with the requested change in camera.
Despite the same camera movement, there is a considerable shift in the position of objects in the generated images due to the scale uncertainty.
This added uncertainty must be represented by GNVS models, potentially wasting model capacity.

The goal of our work is, therefore, to
(a) identify and quantify the issue of scale uncertainty in the GNVS task, and
(b) propose a simple learning-based method for curtailing its impact.
In particular, for (b), we propose to optimize a per-scene scale factor while training the generative model.
As the scales become more consistent, the diffusion objective is easier to optimize as inter-scene inconsistencies shrink.
As a result, our method is able to both reduce the scale uncertainty in the resulting model and improve on short-term reconstruction metrics, and image quality.

\section{Related Work}
\label{sec:relwork}
\textbf{Scale ambiguity.} 
Scene reconstruction via multiview geometry, long-studied in computer vision \cite{hartley2003multiple} and photogrammetry \cite{gosh1981history}, is well known to suffer from scale indeterminacy. For instance, the epipolar constraints between two views are encoded in the essential matrix (e.g., \cite{8-point,5-point}), which can be used to estimate relative camera rotation and translation. However, translation can be estimated only up to an unknown scale factor, as depicted in Fig.~\ref{fig:SA_VO}. This is a central difficulty in precisely calibrating cameras for recovering 3D scene geometry, leading to downstream problems with multiview image datasets, which may not have consistent scales. This issue can also appear within a single scene (the ``relative scale problem'' \cite{cin2023multi}), in the presence of moving objects, which can each only be reconstructed up to a scale.
Similarly, an important task in visual perception for robotics, 
monocular visual odometry (VO), also struggles with scale ambiguity.
Most algorithms (e.g., \cite{Resolving-scale-ambiguity, scale-recovery, eliminating-scale-ambiguity}) use a combination of assumptions and heuristics based on the world ground-plane and camera movements to recover the estimated scale.

The prevalence of machine learning in 3D computer vision necessitates using multiview image datasets.
The scale ambiguity problem, therefore, naturally appears in most real datasets, which use visual methods for camera pose estimation, such as ORBSLAM~\cite{ORBSLAM} and COLMAP~\cite{colmap}. 
Monocular SLAM systems yield consistent camera poses over a sequence, but absolute metric units remain ambiguous.
For instance, the RealEstate10K dataset~\cite{realestate10k} utilizes camera poses and sparse depth estimates recovered for each video clip via ORBSLAM2~\cite{ORBSLAM}. 
To handle the scale inconsistencies across scenes, the authors ``scale-normalized'' each sequence by scaling a specific depth quantile to a reasonable value for their method.
This heuristic works well for their approach but still results in scale inconsistencies between scenes.
In contrast, our method solves for scale values with a learning-based end-to-end approach that encourages consistency between all scenes.

\noindent
\textbf{Generative Novel View Synthesis.} 
Novel view synthesis (NVS) has been a long-standing challenge in computer vision and graphics (e.g., \cite{3D-scene-representation, NVS-in-tensor, View_interpolation}). 
Recent neural 3D scene representations for NVS (e.g., \cite{yu2024mip,barron2023zip}) have been very successful, but 
(i) are fit per-scene, avoiding the scale ambiguity issue, and
(ii) cannot handle unseen or disoccluded scene areas.
To handle such uncertainties, \textit{generative} NVS (GNVS) models have been devised
(e.g., \cite{infinite, lookout, Geofree, synsin}), 
with most recent methods applying 
diffusion models (DMs)~\cite{NVSDiff, GenNVS, photocon, polyoculus, yu2024mip, liu2023zero, liu2023syncdreamer,gao2024cat3d}.
In particular, while some methods, such as GeNVS~\cite{GenNVS}, employ 3D geometric structures,
several are pose-conditional image-based DMs (e.g., \cite{photocon,gao2024cat3d,NVSDiff}), 
which eschew explicit 3D constraints.
In this work, we focus the impact of scale ambiguity in GNVS, building upon a recent state-of-the-art image-based GNVS architecture, PolyOculus~\cite{polyoculus}, 
which can concurrently process conditioning and generated view-sets of any size.
Knowing that the multi-view dataset used to train most GNVS methods (RealEstate10K \cite{realestate10k}) does not contain consistently calibrated scene scales, we show that the model learns to reproduce this unwanted scale variability.
We show that when trained without this noise, the model generates images with much lower implicitly scale variability since many indoor structures are built with consistent dimensions that serve as reliable cues for scale.
This scene scale ambiguity adds an unnecessary challenge for GNVS models, which we address in this paper.

\noindent\textbf{Scale Disambiguation in NVS.}
Some scene-centric NVS methods do attempt to address the scale problem~\cite{zeronvs, SingleViewSynthesis, 4DiM}.
\citet{SingleViewSynthesis} predict an MPI representation from a single view, estimating a scale factor during rendering (based on predicted disparity and SLAM-derived point sets) that reduces scale variability and helps enable NVS training, though not in the generative setting.
 ZeroNVS~\cite{zeronvs} introduces three normalization schemes to address the scale ambiguity issue, all of which rely on statistics from the scene depths and cameras. 
 These normalizations improve generation quality, suggesting the importance of scale normalization.
Recent concurrent work, 4DiM~\cite{4DiM}, which mixes 3D and video data into a 4D generative model, attempts to obtain metric geometry by calibrating the data using a monocular metric depth estimator~\cite{saxena2023zero}.
However, this process is unreliable, leading to ${\sim}30$\% of the data being discarded, whereas our approach is able to utilize all scenes.

In summary, all existing methods, e.g., \cite{SingleViewSynthesis,zeronvs,4DiM} perform scale estimation or normalization as \emph{a 
fixed preprocessing step,} based on approximate heuristics applied to the scene and camera geometries.
As a result mistakes in the initial scales cannot be corrected, valuable training data is potentially sacrificed and the resulting GNVS model learns to reproduce this uncertainty in scale.
In contrast, our approach \textit{learns} a scene scales during the training process, adapting and correcting them via the GNVS task loss.

\noindent\textbf{NVS Metrics.} Existing evaluation metrics for NVS mostly focus on image quality (\eg, FID~\cite{FID2017Heusel}), reconstruction (\eg, LPIPS\cite{FID2017Heusel} and PSNR), geometric consistency (\eg, TSED~\cite{photocon}), and accuracy in camera pose alignment (\eg, SfM distances and SfM Keypoints~\cite{4DiM}).
However, these metrics neither capture the statistical entropy induced by the scale variability nor measure the inconsistencies in scale among the generated views. 
FID only compares distributions of images in an arbitrary feature space with no explicit notion of scale.
Reconstruction-based metrics measure low-level similarity between generated images and a ground truth reference and would suffer due to scale uncertainty however they are quickly overwhelmed by other sources of error (\eg, differences in semantic content, especially further from any observed view).
The SfM-distances and SfM-keypoints metrics, introduced in 4DiM~\cite{4DiM}, use SfM-based methods (\ie COLMAP) and are, as a result, insensitive to scene scale variability. 
Finally, TSED captures geometric consistency between subsequent frames using epipolar geometry which is similarly inherently scale invariant.
To assess the scale variability of a GNVS method, we introduce two new metrics which are directly sensitive to variability in scene scale.
One metric measures the variance of image motion in generated images given a fixed camera motion, while the other, based on TSED, measures the geometric consistency of generation in two independent directions.
We use these metrics to both establish the problem of scale uncertainty in GNVS and measure the efficacy of our proposed approach.

\section{Methods}
\label{sec:method}
GNVS uses a generative model to synthesize novel views of an object or scene, given limited observations.
In the simplest case, given one observed view while generating one novel view, GNVS methods seek to model $p(\mathbf{x} | \mathbf{x}_o, \mathbf{c})$, where $\mathbf{x}$ is the novel view conditioned on an observed view $\mathbf{x}_o$, and the relative camera information, $\mathbf{c} = (\mathbf{K},\mathbf{R},\mathbf{t})$, which contains camera intrinsics, $\mathbf{K}$, and extrinsics, including rotation, $\mathbf{R}$, and translation, $\mathbf{t}$.
Although the underlying scene is 3D in nature, this formulation treats NVS as an image-to-image mapping problem, and all factors of variation are implied in the images.
When synthesizing a novel view in this minimal setting, the model must implicitly ``select'' a scene scale that explains the apparent motion in the generated view.
Ideally, by disentangling this scale factor from the data, we reduce the space in which the noise manifests from the high-dimensional data space to a single scaler.
Without this, the scale ambiguity potentially contaminates GNVS training with artificially induced data variability. 
On the other hand, if a consistent scale is used, it makes the denoising task easier by reducing the space of possible new views to those with a single consistent scale.
We therefore argue that \textit{optimization of the scales is aligned with the diffusion objective used in GNVS}. 
Hence, by treating the per-scene scales as a parameter during training, consistent scale factors can be estimated.

\subsection{Scale Learning}
Let $S = \{{s_i}\}_{i=1}^N$ represent the per-scene scales for $N$ scenes, which we aim to learn.
These scene scales modify the camera extrinsics for each frame in a scene by scaling the translation, i.e., $\hat{\mathbf{c}}_j = (\mathbf{K}_j,\mathbf{R}_j, \hat{\mathbf{t}}_j)$, where $\hat{\mathbf{t}}_j = s_{i} \mathbf{t}_j$ is the translation scaled by the scale parameter of scene $i$, to which frame $j$ belongs.
We assume the base scales implied by the camera translations, $\mathbf{t}_j$, obtained by whichever calibration and normalization procedure was applied, are a reasonable initialization.
For optimization, the scales are parameterized as 
\begin{equation}
    s_i = \exp(a [\beta_i]_{-1}^{+1}),
\end{equation} where $\beta_i$ are learnable parameters, $[z]_{-1}^{+1} = \mathbf{Clamp}(z, -1, +1)$, and the parameter $a$ is a hyperparameter which controls how large or small scales can become.
The exponential ensures that scales remain strictly positive, and the use of the clamping prevents scales from diverging.
We set $a=1$, equivalent to assuming that the base scales are within a factor of $e$ of a consistent scale.
\begin{figure*}[t]
    \centering
    \includegraphics[width=0.99\textwidth]{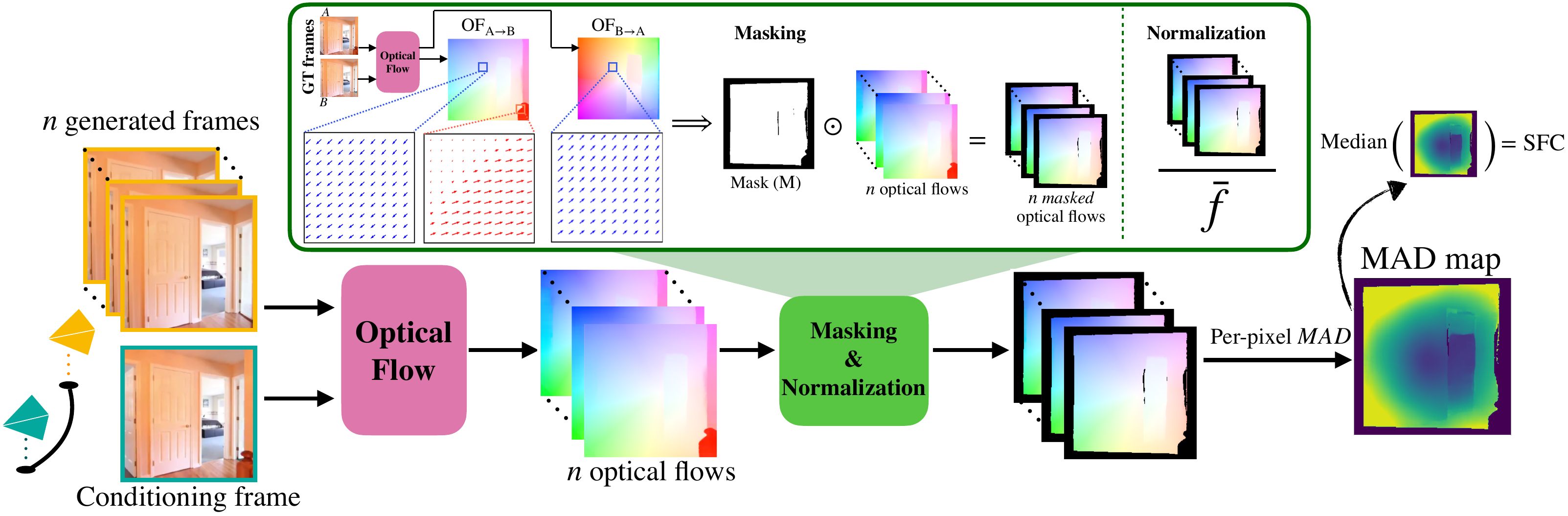}
    \caption{
    \textbf{SFC metric overview.} Given a set of size $n$ of generated frames, we use an optical flow estimator to obtain optical flows between the conditioning and generated frames.
    The flows are masked and normalized. To illustrate the mask generation process, optical flow vectors of two patches are shown: a masked patch and an unmasked patch. Red flow vectors represent a group of masked pixels moving out of the field-of-view by forward flow. Blue flow vectors correspond to a group of unmasked pixels whose flows are in opposite directions, indicating cycle consistency. After masking, we normalize the masked flows with the average flow magnitude of unmasked pixels, $\bar{f}$.
    Finally, we compute per-pixel median absolute deviation (MAD) over $n$ masked normalized flows to get the MAD map, the median value of which is defined as \textit{SFC}.
    }
    \label{fig:sfc_fig}
\end{figure*}

We learn the scales, $S$, jointly with the parameters, $\theta$, of the diffusion model using a standard diffusion loss\ \cite{ddpm}.
Here, we use PolyOculus\ \cite{polyoculus}, a GNVS model based on latent diffusion process\ \cite{rombach2022high}.
Specifically, we use
\begin{align}
\label{eq:loss_method}
\mathcal{L} = \vert \vert \epsilon - \epsilon_{\theta}(\{(\mathbf{z}_{1}, t_{1}, \hat{\mathbf{c}}_{1}), . . . , (\mathbf{z}_{K}, t_{K}, \hat{\mathbf{c}}_{K})\})\vert \vert ^2, 
\end{align}
where $K \in [1,5]$ is the uniformly selected number of total frames, $z_{k}$ is the latent encoding for the $k^{th}$ image, $\hat{\mathbf{c}}_{k}$ is scaled camera poses of frame $k$, and $\epsilon_\theta$ is the denoiser which takes in all the inputs including the camera poses and estimates the added noise, $\epsilon$.
From the $K$ views, $[0,k-1]$ of the views are uniformly selected as conditioning, where for those images, $t_k$ is $0$ and no noise is added during training.

\subsection{Quantifying Scale Variability}

We define two new metrics which directly measure the variability of scale learned by a GNVS method.
The first metric derives from the observation that, when looking at multiple generated images conditioned on the same image and camera pose, the observed motion between the conditioning and generated images should be consistent.
Intuitively, we can measure this motion by computing optical flow between the conditioning image and generated images and quantifying its consistency through the variance of those flow fields, and thus, we call this \textit{Sample Flow Consistency (SFC)}.
The second metric is based on the observation that if we generate two images with different camera poses while conditioned on the same starting point, the epipolar geometry between the two generated images is known and should be respected as if they were both generated with a consistent scale.
We measure this by computing TSED \cite{photocon} between two such generated images and call this the \textit{Scale-Sensitive Thresholded Symmetric Epipolar Distance (SS-TSED)}.
We describe these two metrics in more detail next.

\paragraph*{Sample Flow Consistency.}
An overview of the SFC metric is given in Fig.~\ref{fig:sfc_fig}. 
For a given condition image, we generate a set of size $n$ images conditioned on the same frame with the same camera motion between the conditioned and generated frames.
We compute the forward optical flow, $f_i$, from the conditioned image and the $i$th generated image.
To avoid using parts of the image that have inconsistent flow or become occluded or disoccluded, we compute a mask $M$ of the image by checking for cycle consistency\ \cite{ChangCC21, BlackA96, SundaramBK10} of the optical flow between the conditioned frame and the ground-truth frame for the specified camera motion.
For cases where the ground truth is unavailable, we instead compute a consensus mask based on cycle consistency of the flow between the conditioned frame and the generated frames.
Here, we focus primarily on computing SFC on test sequences with the test camera motions and do not need to do this.
For convenience, we use RAFT \cite{raft} for the optical flow computation, though other choices are possible.

The magnitude of observed image motion varies based on factors other than just scale, including the geometric distribution of objects in a scene (i.e., nearby objects tend to have larger motions, leading to larger flows) and the magnitude of the camera motion itself.
To avoid our metric being dominated by scenes with closer content or larger camera motions, we normalize the optical flow for each scene before estimating its variance. 
Specifically, we compute the average magnitude, $\bar{f}$, of the forward optical flow at unmasked pixels with the generated images.
We then normalize the forward flows as $\hat{f}_i = f_i / \bar{f}$.

\begin{figure}[t]
   \includegraphics[width=\linewidth]{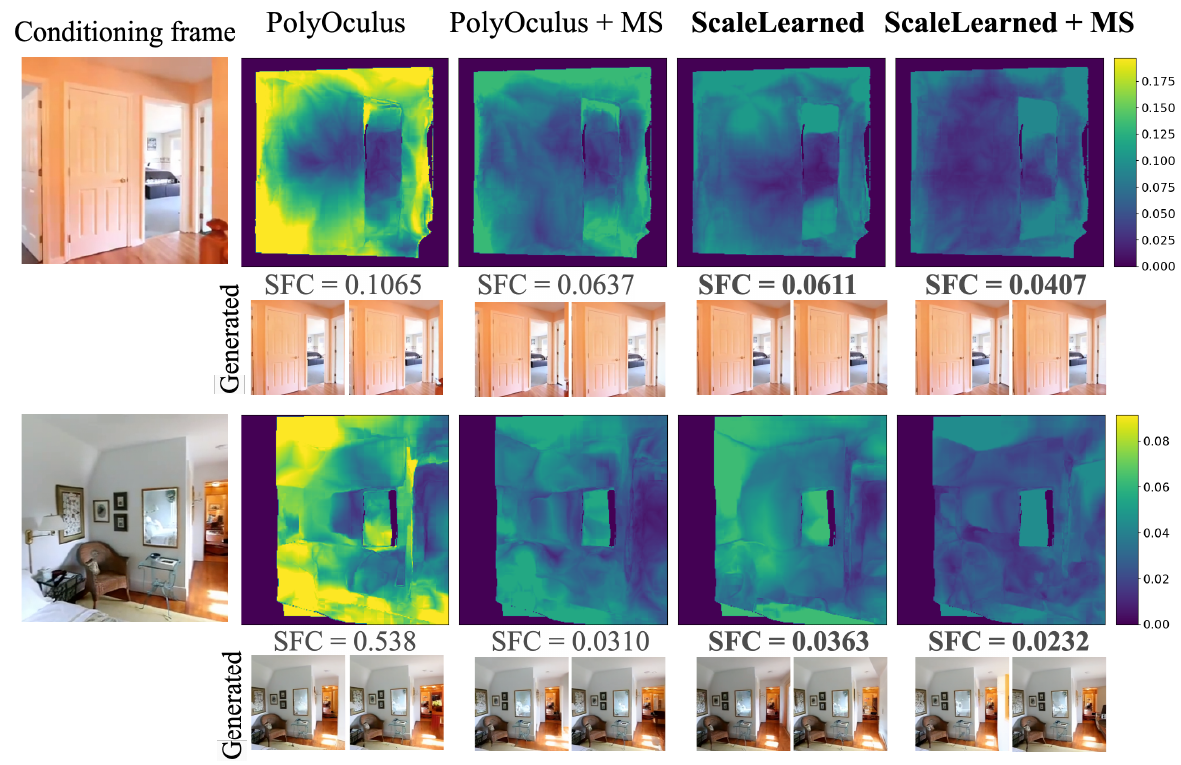}
   \caption{\textbf{Examples of per-pixel optical flow MAD maps.} The darker the pixel, the lower the variation of optical flow in that pixel, indicating a more consistent scale among the generated samples. The entropy in scale can also be seen by comparing the generated frames, \eg, the width of the door in the second row.}
   \label{fig:sfc_heatmaps}
\end{figure}

Finally, we seek to compute the variance of the normalized flow between generated samples.
To be robust to outliers in optical flow, we use the per-pixel median absolute deviation (MAD) across all $n$ masked normalized flows.
Specifically, let
\begin{equation}
    \textrm{MAD}[p] = \textrm{median}_i \left| \hat{f}_i[p] - \textrm{mean}_j \hat{f}_j[p] \right|
\end{equation}
be the median average deviation at pixel, $p$, where the median and mean are both taken over a set of $n$ generated images where pixel $p$ is unmasked (i.e., $M_i[p] = 1$).
The SFC of a single conditioning image and desired camera pose,
\begin{equation}
    \textrm{SFC} = \textrm{median}_{p \in M_*} \textrm{MAD}[p],
\end{equation}
where the median is taken over pixel locations, $p$, and $M_*$ is the cycle consistency mask computed between the conditioning image and the ground truth novel view.
Note that $M_*$ is only available when the ground truth novel view is available, e.g., when using test set camera trajectories.  When it is unavailable, we instead use $M_* = (\textrm{mean}_i M_i) > \epsilon$.
We average the $\textrm{SFC}$ of different conditioning images and camera poses to create a final metric.
We visualize the $\textrm{MAD}$ maps as heatmaps and show SFC values for different scenes in Fig.\ \ref{fig:sfc_heatmaps}.

To summarize, the SFC metric quantitatively measures the scale entropy of the samples of the same scene with a fixed camera motion. Smaller SFC values indicate less variation in the generated samples, suggesting more scale consistency.

\begin{figure}[t]
    \centering
    \includegraphics[width=0.8\linewidth]{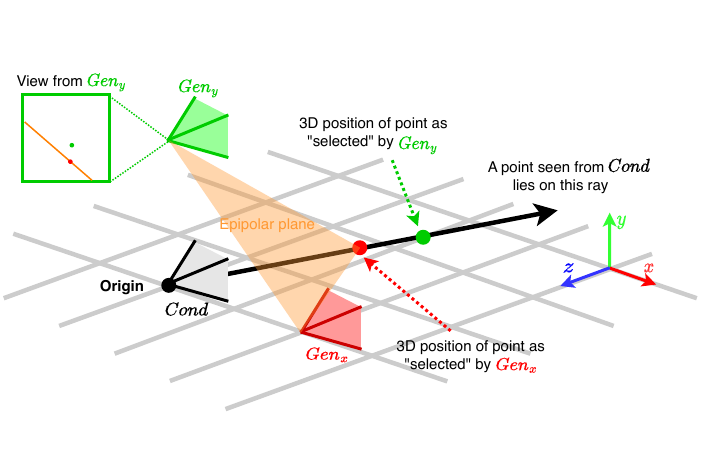}
    \caption{ \textbf{SS-TSED visualization.}
    A visualization of using TSED between two generated views, $Gen_x$ and $Gen_y$, to measure scale variability.
    Each generated view, with different scene scales, results in the 3D position of a point observed in $Cond$ to be different ({\color{red}red} and {\color{green}green} dots).
    A difference in scale causes the 2D position of the point observed from $Gen_y$ to lie a distance off the {\color{orange}epipolar line/plane} formed by the generated views and the point observed by $Gen_x$.
    }
    \label{fig:tsed_setup}
\end{figure}

\paragraph*{Scale-Sensitive Thresholded Symmetric Epipolar Distance.}
By design, TSED is insensitive to changes in scene scale between a generated and conditioning view.
A point in the conditioning view lies along a ray originating from that view.
By generating a view from a different perspective, that point becomes constrained to a unique position along that ray.
Since the epipolar geometry is defined in essence by the ray and camera centers, the point will always lie along the epipolar plane for different scene scales.

Instead of measuring TSED between a generated and conditioning view, the TSED between two different novel views conditioning on the same view can be sensitive to the different scene scales ``chosen'' in each generation.
A point will always lie on a ray originating from the conditioning view, but different scene scales from each generated view will place the point at different distances along the ray.
If the poses of the generated views are chosen correctly, inconsistencies in the scene scale will place the 2D position of the point from one view some distance off the epipolar plane defined using the 3D position of the point from the other view.
An illustration of this is shown in Fig.\ \ref{fig:tsed_setup}.

To compute this metric for a single conditioning image, for each axis (x-, y-, and z-), we randomly generate a camera translation along that axis of a fixed magnitude and in a random direction (positive or negative) and then generate an image with that camera motion.
Only camera translations are considered since rotations are unaffected by scene scale.
We sample pairs of such generated images with different axes and then compute the percentage of consistent pairs of images using TSED with a constant $T_\text{matches}=10$ and a range of values for $T_\text{error}$; see \cite{photocon} for more on TSED.
This is then repeated for many images and averaged to arrive at the \textit{Scale Sensitive Thresholded Symmetric Epipolar Distance}, or SS-TSED score.

\begin{figure*}[tbh]
    \begin{subfigure}[b]{0.19\textwidth}
         \centering
         
         \includegraphics[width=\linewidth]{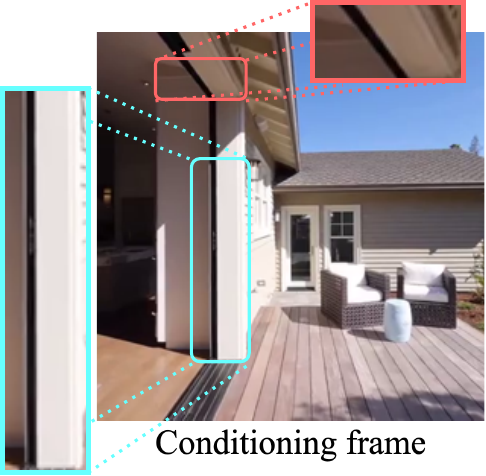}
    \end{subfigure}
    \begin{subfigure}[b]{0.19\textwidth}
         \centering
         
         \includegraphics[width=\linewidth]{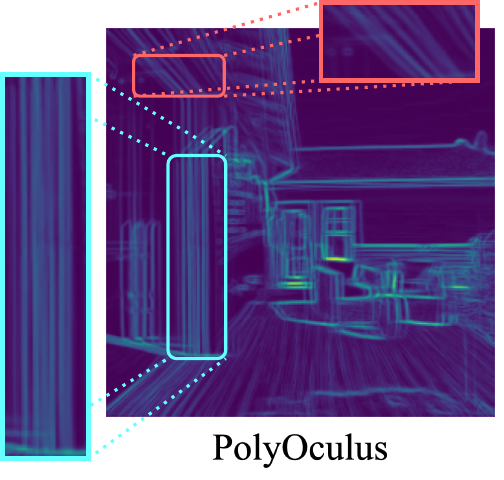}
    \end{subfigure}
    \begin{subfigure}[b]{0.19\textwidth}
         \centering
         \includegraphics[width=\linewidth]{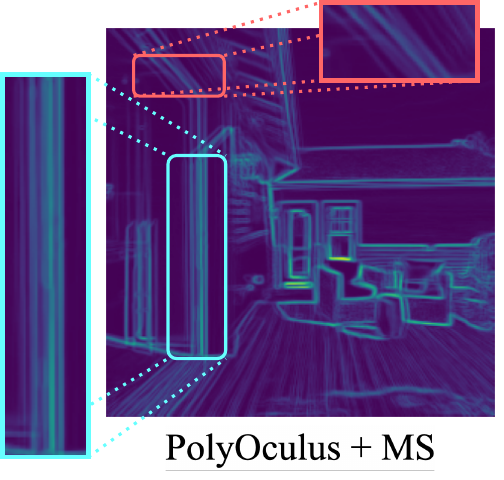}
     \end{subfigure}
     \begin{subfigure}[b]{0.19\textwidth}
         \centering
         \includegraphics[width=\linewidth]{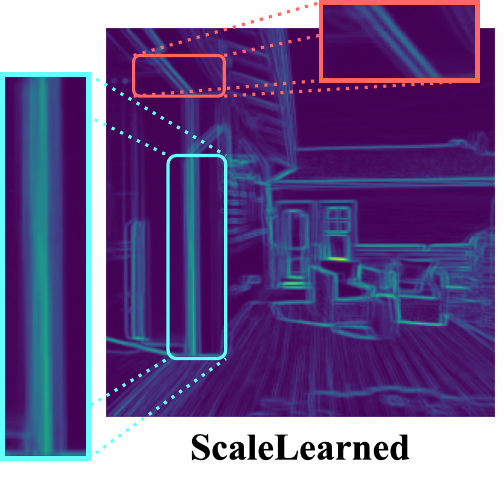}
     \end{subfigure}
     \begin{subfigure}[b]{0.19\textwidth}
         \centering
         \includegraphics[width=\linewidth]{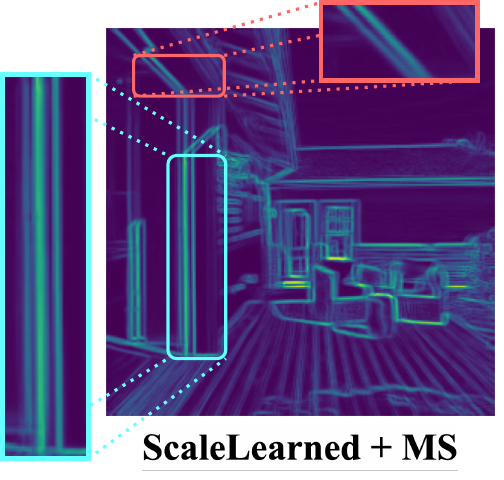}
     \end{subfigure}
    \caption{
        \textbf{Edge heatmaps.}
        We visualize the average Sobel filter responses of multiple samples generated with the same conditioning information to highlight distinct regions of the scene structure. Note that PolyOculus samples are quite noisy in terms of edge locations, and using metric-depth scales helps stabilize edge locations.
        Such consistency in edge locations results in more clarity in the edge heatmaps, which indicates a reduction in randomness caused by scale ambiguity.
        Additional visualizations of scene scale variability are provided in the supplement.
        }
    \label{fig:edge_maps}
\end{figure*}

\section{Experiments}
\label{sec:experiments}
Evaluation of our scale learning method is challenging since datasets are uncalibrated for both training and testing splits.
Hence, we explore a variety of methods to evaluate the reduction of scale uncertainty.
First, we use SFC and SS-TSED as metrics for evaluating the amount of scale variability a model exhibits in its generations.
Next, we evaluate the effect of scale learning on image quality in GNVS using both FID and reconstruction metrics.
Notably, reconstruction metrics on test scenes pose a unique challenge as their scales are not estimated during training, but we show that our approach can be used to estimate them as well.
Finally, we compare our scale learning approach to the use of estimated scales based on monocular metric depth methods, as was recently proposed \cite{4DiM}.

\subsection{Experimental Setup}
\textbf{Datasets.} We evaluate on the RealEstate10K (RE10K)~\cite{realestate10k} dataset, a standard dataset used in GNVS.
We center-crop and downsample the videos to a resolution of $256 \times 256$.
RE10K's videos are segmented into disjoint sequences, where the provided camera parameters were estimated using ORBSLAM2 \cite{ORBSLAM} and ``normalized'' using an ad-hoc method~\cite{realestate10k}.
In addition to the raw camera poses provided by the dataset, we also consider poses calibrated using metric monocular depth estimators (MS) as a representative method for existing ad-hoc scale estimation methods.
We set the scale to $1.0$ for scenes with unreliable scale estimates, resulting in ${\sim}30\%$ of the metric scales staying the same as those provided by RE10K.
4DiM~\cite{4DiM} instead removed scenes from the training set where metric scale estimation failed; here, we compare both trimming these scenes from the dataset and keeping them.
We use both sets of poses to act as \textit{reference scales} from which we apply our scale learning.

\noindent\textbf{GNVS Model.} We use PolyOculus~\cite{polyoculus} as our baseline GNVS model.
PolyOculus is a state-of-the-art multi-view latent diffusion model with a U-Net~\cite{unet} denoising backbone and a VQ-GAN autoencoder~\cite{vqgan}.
To compare the effect of scale learning, we train two models for each reference scale, one with and one without scale learning.
All models are trained from scratch for $900$ epochs. %
For models using scale learning, we initialize the learnable parameters $\beta_i$ to zero, equivalent to initializing all scales to be $1.0$.
We use different optimizers for the scale factors and the denoising network.
We employ the AdamW optimizer~\cite{adamw} with a learning rate of $2.0 \times 10^{-6}$ for the U-Net parameters, like PolyOculus~\cite{polyoculus}, and the Adam optimizer with a learning rate of $10^{-4}$ for the scale parameters.
Training is performed using eight NVIDIA L40 GPUs with a total batch size of 32.

\subsection{Scale Variability as Apparent Motion}
As previously observed, a model trained with inconsistent scales produces novel views with inconsistent apparent motion despite sampling with the same conditioning.
Here, we evaluate the impact of our scale learning approach on this motion both qualitatively and quantitatively using our proposed SFC metric.
We compute SFC as described previously using 200 randomly selected test images.
For each image, we select poses at various distances from the observed view along ground-truth camera trajectories to evaluate scale variation at multiple magnitudes of camera motion.
Without this stratification, there may be additional variance introduced into the samples due to differences in framerate or general camera motion from the dataset trajectories.
Specifically, we select views with distances that match the closest to this set of magnitudes: $T = \{0.05, \allowbreak 0.1, \allowbreak 0.15, \allowbreak 0.2, \allowbreak 0.25, \allowbreak 0.3\}$.
Generally, the magnitudes chosen are relatively small to avoid introducing additional variance when generating new scene content in unobserved regions, which would reduce the usefulness of the computed optical flow.
Finally, for each generated and observed view, we draw 10 samples to evaluate the scale uncertainty produced by the model.

\begin{figure}[t]
    \centering
    \begin{subfigure}{\linewidth}
        \pgfplotsset{width=\linewidth,height=4.5cm,compat=1.18}
        \begin{tikzpicture}
            \begin{axis}[
                title style={align=center, font=\small},
                title={SFC on test-set trajectories $\downarrow$},
                xlabel={Translation magnitude},
                ylabel={Average SFC},
                xmin=0.04, xmax=0.31,
                ymin=0.04, ymax=0.16,
                xtick={0.05,0.10,0.15,0.20,0.25,0.30},
                ytick={0.06,0.08,0.10,0.12},
                legend pos=north east,
                legend style={nodes={scale=0.4, transform shape}},
                label style={font=\scriptsize},
                tick label style={font=\tiny, /pgf/number format/fixed, /pgf/number format/precision=2},
                ymajorgrids=true,
                grid=none,
                xlabel style={yshift=0.5ex},
                ylabel style={yshift=-1ex},
                mark size=1.0pt,
                scaled ticks=false
            ]
                \addplot[color=blue,mark=*,dashed] coordinates {
                    (0.05,0.1321733)(0.1,0.1116090)(0.15,0.10325491)(0.2,0.09853840)(0.25,0.09291527)(0.3,0.08963124)
                };
                \addplot[color=orange,mark=square*,dashed] coordinates {
                    (0.05,0.1318924)(0.1,0.1021049)(0.15,0.09409188)(0.2,0.08832081)(0.25,0.08416181)(0.3,0.07705204)
                };
                \addplot[color=blue,mark=*,] coordinates {
                    (0.05,0.0918330)(0.1,0.0739316)(0.15,0.06384188)(0.2,0.05823862)(0.25,0.05492046)(0.3,0.05033673)
                };
                \addplot[color=orange,mark=square*,] coordinates {
                    (0.05,0.0925231)(0.1,0.0719607)(0.15,0.05927127)(0.2,0.05590342)(0.25,0.05128476)(0.3,0.04737328)
                };
                \legend{
                    PolyOculus,
                    PolyOculus (with MS),
                    ScaleLearned,
                    ScaleLearned (with MS),
                }
            \end{axis}
        \end{tikzpicture}   
        \label{fig:comb_sfc_orig}
    \end{subfigure}
    \caption{
     \textbf{SFC evaluations.} We report average SFC values for each translation amount over 200 randomly selected test scenes for test set camera poses.  While the use of metric scales alone does provide some improvement in SFC (PolyOculus with MS), scale learning provides a more substantial improvement.  The combination of metric scales with scale learning (ScaleLearned with MS) provides a small further improvement in SFC.
    }
    \label{fig:sfc_combined}
\end{figure}
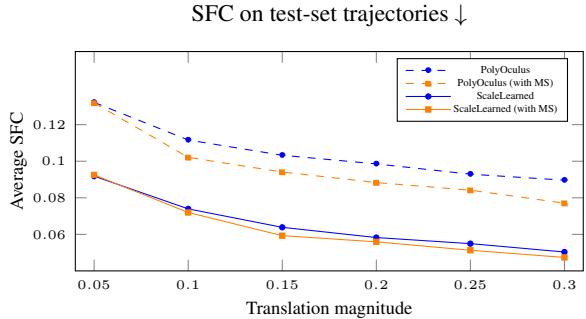

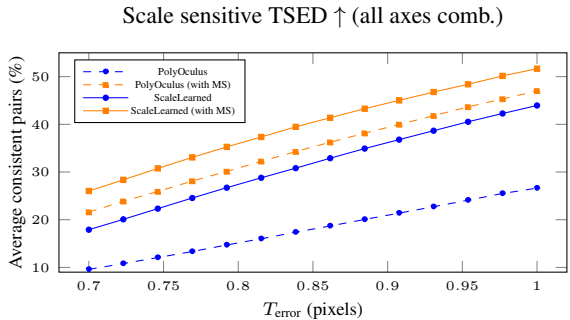
\begin{figure}[t]
  \centering
    \pgfplotsset{width=\linewidth,height=4.5cm,compat=1.18}
    \begin{tikzpicture}
        \begin{axis}[
            title style={align=center, font=\small},
            title={Scale sensitive TSED $\uparrow$ (all axes comb.)},
            xlabel={$T_\text{error}$ (pixels)},
            ylabel={Average consistent pairs (\%)},
            xmin=0.68, xmax=1.02,
            ymin=9, ymax=55,
            xtick={0.7,0.75,0.8,0.85,0.9,0.95,1.0},
            ytick={10,20,30,40,50},
            legend pos=north west,
            legend style={nodes={scale=0.4, transform shape}},
            label style={font=\scriptsize},
            tick label style={font=\tiny, /pgf/number format/fixed, /pgf/number format/precision=2},
            ymajorgrids=true,
            grid=none,
            xlabel style={yshift=0.5ex},
            ylabel style={yshift=-1ex},
            mark size=1.0pt,
            scaled ticks=false
        ]
            \addplot[color=blue,mark=*,dashed] coordinates {
                (0.7,9.583243145743147)(0.7230769230769231,10.805904077090515)(0.7461538461538462,12.062047533445835)(0.7692307692307692,13.324539278920634)(0.7923076923076923,14.69859001638209)(0.8153846153846154,16.00367323337662)(0.8384615384615385,17.381476691396912)(0.8615384615384616,18.686882444297694)(0.8846153846153846,20.052308802308648)(0.9076923076923077,21.38077047716878)(0.9307692307692308,22.72495383618264)(0.9538461538461538,24.090013329420106)(0.976923076923077,25.498755716973175)(1.0,26.62512993113275)
            };
            \addplot[color=orange,mark=square*,dashed] coordinates {
                (0.7,21.59859992441663)(0.7230769230769231,23.861218503271477)(0.7461538461538462,25.90713362368992)(0.7692307692307692,28.111882560706402)(0.7923076923076923,30.122918779684635)(0.8153846153846154,32.23687628014265)(0.8384615384615385,34.271633554083884)(0.8615384615384616,36.25004822703499)(0.8846153846153846,38.138544339018786)(0.9076923076923077,39.97638665155221)(0.9307692307692308,41.80561521786685)(0.9538461538461538,43.65459002048405)(0.976923076923077,45.31572349593068)(1.0,47.02853200883002)
            };
            \addplot[color=blue,mark=*,] coordinates {
                (0.7,17.90260742839165)(0.7230769230769231,20.074043877269684)(0.7461538461538462,22.32386545354287)(0.7692307692307692,24.548838949827836)(0.7923076923076923,26.717027058962532)(0.8153846153846154,28.788892827602506)(0.8384615384615385,30.807664736410256)(0.8615384615384616,32.889708929063765)(0.8846153846153846,34.91600141792579)(0.9076923076923077,36.794540942928045)(0.9307692307692308,38.66164795777699)(0.9538461538461538,40.53623616526843)(0.976923076923077,42.2727433025641)(1.0,43.94889991506479)
            };
            \addplot[color=orange,mark=square*,] coordinates {
                (0.7,26.02297002329779)(0.7230769230769231,28.3569659534848)(0.7461538461538462,30.762644244152)(0.7692307692307692,33.06750489820323)(0.7923076923076923,35.273045468915086)(0.8153846153846154,37.34793666736352)(0.8384615384615385,39.46971568605679)(0.8615384615384616,41.378599936648705)(0.8846153846153846,43.28139823939846)(0.9076923076923077,45.024482487209745)(0.9307692307692308,46.77722425394263)(0.9538461538461538,48.3915207187491)(0.976923076923077,50.15527447420976)(1.0,51.66433840954704)
            };
            \legend{
                PolyOculus,
                PolyOculus (with MS),
                ScaleLearned,
                ScaleLearned (with MS),
            }
        \end{axis}
    \end{tikzpicture}   
   
   \caption{\textbf{SS-TSED evaluation.}  Consistent image pairs percentage computed with TSED on sample pairs with camera translations in different directions. The x-axis is $T_{error}$, and the y-axis shows the proportion of pairs (see supplement for separate plots of each combination.)
   We again see that scale learning provides a meaningful improvement in consistency with a further improvement with the use of metric depth-based scales.
   }
   \label{fig:tsed}
\end{figure}

The effect of scale inconsistency is observable via qualitative inspection of the novel views as different amounts of apparent motion in each novel view.
The variance in motion among novel views with the same generation parameters can be made more salient by averaging Sobel filter~\cite{vision-text-book} outputs from multiple samples to form heatmaps, which are shown in Fig.~\ref{fig:edge_maps}.
Consistent scales should produce consistent motion and lead to sharp edges in the heatmaps, while variability in scale leads to blurred or ghosted edge maps.
In Fig.~\ref{fig:edge_maps}, we see that, indeed, the baseline PolyOculus model with fixed scales produces samples with large amounts of scale variability, where magnified regions show edges that become blurred, appearing in multiple locations between the samples.
This holds true even when using the metric depth-based reference scales in PolyOculus+MS. 
In contrast, scale learned models show sharper heatmaps.
This can also be observed when inspecting MAD heatmaps in Figure\ \ref{fig:sfc_heatmaps}, where the ScaleLearned models produce darker maps and lower SFCs due to lower per-pixel MAD, reflecting reduced optical flow variation.

We can also quantify these improvements more precisely using SFC.
As shown in Figure~\ref{fig:sfc_combined}, models incorporating scale learning outperform those without it in terms of SFC across a range of translation magnitudes, demonstrating that learning scale during training reduces motion variation between the conditioning and generated frames.
Using metric-depth scales as reference scales (ScaleLearned + MS) further enhances scale consistency, as they are less noisy compared to the dataset scales and more closely aligned with actual scene scales.
Although metric-depth scaled camera poses help reduce sample scale variability, ScaleLearned models outperform it by a significant margin, suggesting the scale-learning method to be more useful compared to metric-depth estimated scales.
While metric-depth scaled camera poses help reduce sample scale variance (PolyOculus + MS), ScaleLearned models achieve significantly better performance, indicating that the scale-learning approach is more effective than metric-depth estimation in preserving scale consistency.
Further improvements in scale consistency are observed when using metric-depth scales as reference scales in scale learning (ScaleLearned + MS), as they exhibit lower noise compared to dataset scales and better align with actual scene scales.
This is also supported by qualitative inspections of the MAD heatmaps in Fig.~\ref{fig:sfc_heatmaps}, where the ScaleLearned models with darker maps correspond with lower SFCs due to lower per-pixel MAD, reflecting lower optical flow variation.

\subsection{Scale Variability as Epipolar Errors}
We compute SS-TSED over 200 images in the RE10K test set and for each image, we sample 100 pairs.
The results in Fig.~\ref{fig:tsed} demonstrate that incorporating scale learning significantly improves epipolar consistency across both reference scale cases.
Additionally, using metric-depth estimated scales enhances TSED, and further performance gains are achieved by combining scale learning with this approach.

\subsection{Scale Variability as Reconstruction Errors}
As mentioned in Section \ref{sec:relwork}, reconstruction-based metrics are not designed to measure scale variations. 
Reconstruction errors in GNVS can generally be attributed to two factors: mismatches due to the generation of new content in unseen regions and the misalignment of scene content due to different scene scales.
Although reconstruction errors due to the generation of new content can be mitigated by using small camera motions, errors due to content misalignment can still occur.
Since we do not learn the scale of test scenes, content in novel views may be consistently misaligned due to a difference in scale between the ground-truth scene and the trained model.

To mitigate the issue and make reconstruction metrics more sensitive to scale variability, we propose a test-time scale estimation procedure.
Specifically, we can use the same diffusion loss as used in training but freeze the model weights and \textit{only} learn the scales of a new scene.
We use this approach to estimate scales for 1500 random test scenes using our scale learning method.
Metric-depth reference scales for the test-set are used for models that also use them for training, and are estimated using the same method.
For fairness, we also apply test-time scale estimation even for models that do not use scale learning during training.
This is performed separately per model since calibration only resolves scale relative to a model's internal notion of scale, which may vary by model.
The same trend found in our previous experiments is also found here in Table\ \ref{tab:ref_quality}.
Scale learning consistently improves reconstruction, and performs slightly better when applied in conjunction with metric-depth calibrated reference scales.
\begin{table}
    \centering
    \footnotesize
    \resizebox{\linewidth}{!}{
        \begin{tabular}{l|cccc|cccc}
            \multicolumn{1}{c}{\multirow{2}{*}{Method}}  & \multicolumn{4}{c}{LPIPS $\downarrow$} & \multicolumn{4}{c}{PSNR $\uparrow$}\\
            & 1 & 2 & 3 & 4 & 1 & 2 & 3 & 4 \\
            \hline
            PolyOculus& 0.037 & 0.045 &  0.053 &  0.061 &  29.98 & 27.38 & 26.32 & 25.21\\       
            PolyOculus + MS & 0.037 & 0.046 & 0.052 &  0.060 & 29.10 & 27.61 & 26.56 & 25.56\\
            ScaleLearned & \underline{0.035} & \underline{0.042} & \underline{0.049}& \underline{0.056} & \underline{29.50} & \underline{28.13} & \underline{27.17}& \underline{26.17} \\
            ScaleLearned + MS & \textbf{0.034} & \textbf{0.041} & \textbf{0.048} & \textbf{0.055} & \textbf{29.62} & \textbf{28.22} & \textbf{27.24} &  \textbf{26.24}\\
        \end{tabular}
    }%
    \caption{
    \textbf{Reconstruction metrics.} Computed on the test set for one to four frames ahead.  Scale learning shows improvements in reconstrunction metrics.}
    \label{tab:ref_quality}
\end{table}

\subsection{Trimming Unreliable Scenes}
\label{sec:results-trimming}
In 4DiM~\cite{4DiM}, training scenes with inconsistent calibrations were shown to degrade performance in terms of image fidelity and scale alignment, where it was proposed to use metric depth-based scale estimation.
Unfortunately, this procedure is not always successful, as we noted above, and \citet{4DiM} proposed simply dropping these scenes from the training set in the creation of the Calibrated RE10K (cRE10K) dataset, resulting in a ${\sim}30\%$ reduction in training data.
This has the potential upside of both making scales more consistent and scale learning easier at the cost of losing some amount of data.
In contrast, in our experiments with metric-depth-based scales, the scale of unreliable scenes was initialized to 1 (i.e., left at the default dataset scale) but subsequently learned, allowing us to keep all data but potentially complicating training.
Here, we compare these two approaches.
In this experiment, we perform evaluations using the same settings previously described with SFC, and on long trajectory generation where image quality is measured with FID. 

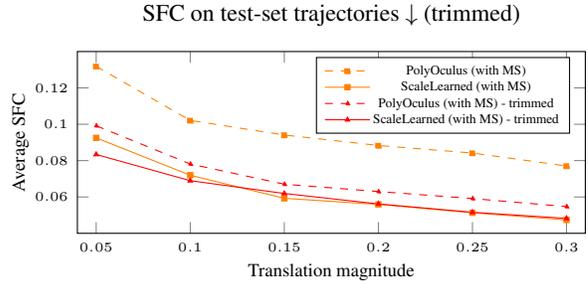
\begin{figure}[t]
  \centering
   \pgfplotsset{width=\linewidth,height=4cm,compat=1.18}
    \begin{tikzpicture}
        \begin{axis}[
            title style={align=center, font=\small},
            title={SFC on test-set trajectories $\downarrow$ (trimmed)},
            xlabel={Translation magnitude},
            ylabel={Average SFC},
            xmin=0.04, xmax=0.31,
            ymin=0.04, ymax=0.14,
            xtick={0.05,0.10,0.15,0.20,0.25,0.30},
            ytick={0.06,0.08,0.10,0.12},
            legend pos=north east,
            legend style={nodes={scale=0.5, transform shape}},
            label style={font=\scriptsize},
            tick label style={font=\tiny, /pgf/number format/fixed, /pgf/number format/precision=2},
            ymajorgrids=true,
            grid=none,
            xlabel style={yshift=0.5ex},
            ylabel style={yshift=-1ex},
            mark size=1.0pt,
            scaled ticks=false
        ]
            \addplot[color=orange,mark=square*,dashed] coordinates {
                (0.05,0.13189243)(0.1,0.10210497)(0.15,0.09409188)(0.2,0.08832081)(0.25,0.08416181)(0.3,0.07705204)
            };
            \addplot[color=orange,mark=square*,] coordinates {
                (0.05,0.09252317)(0.1,0.07196076)(0.15,0.05927127)(0.2,0.05590342)(0.25,0.05128476)(0.3,0.04737328)
            };
            \addplot[color=red,mark=triangle*,dashed] coordinates {
                (0.05,0.09917963)(0.1,0.07814313)(0.15,0.06702464)(0.2,0.06299425)(0.25,0.05913413)(0.3,0.05470605)
            };
            \addplot[color=red,mark=triangle*,] coordinates {
                (0.05,0.08337945)(0.1,0.06895573)(0.15,0.06192213)(0.2,0.05617108)(0.25,0.05157790)(0.3,0.04821511)
            };
            \legend{
                PolyOculus (with MS),
                ScaleLearned (with MS),
                PolyOculus (with MS) - trimmed,
                ScaleLearned (with MS) - trimmed,
            }
        \end{axis}
    \end{tikzpicture}
   \caption{
   \textbf{Effects of trimming vs scale learning on SFC.}  Scale learning has en effect on SFC comparable to trimming unreliably calibrated scenes entirely but without the loss of data.
   }
   \label{fig:sfc_trimmed}
\end{figure}

Figure\ \ref{fig:sfc_trimmed} shows that trimming reduces scale variability for all methods, where the largest improvements are gained when no scale learning is used.
Trimming only yields small improvements for the ScaleLearned + MS model, suggesting that the scale learning is able to effectively compensate for scenes where metric scale estimation failed.

However, trimming throws away a non-trivial fraction of the dataset, which may otherwise hurt the quality of the resulting model.
Thus, we evaluate using FID for long-term trajectory generation, where we report the FID on the final frames where image quality degrades the most due to error accumulation~\cite{polyoculus}.
Table\ \ref{tab:keyframe-FID} shows trimming actually hurts image quality in terms of FID, which should not be surprising as it has substantially reduced the amount of training data available.
These results further show that the use of metric scales on their own, without trimming or scale learning, is harmful to image quality, likely due to outliers from the metric calibration process that cannot be compensated for during training.
Finally, we see that scale learning improves over the baseline method with neither trimming nor metric reference scales, suggesting that the reduction of scale inconsistency during training allows the model to more effectively learn the conditional image distribution.

\begin{table}
    \centering
    \resizebox{\linewidth}{!}{
    \footnotesize
    \begin{tabular}{ccc|ccc}
        \multicolumn{3}{c}{} & \multicolumn{3}{c}{FID $\downarrow$} \\
        \makecell{Reference \\scale} & \makecell{Scale \\learning} & Trim & Last 3 & Last 5 & All\\
        \hline
         Dataset      & \xmark & --     & 34.82 & 34.21 & 29.02 \\
         Metric-depth & \xmark & \xmark &  37.35 & 36.52 & 30.58\\
         Metric-depth & \xmark & \cmark & 35.28 & 34.64 & 29.46\\
         Dataset      & \cmark & -- & \textbf{33.12} & \textbf{32.61} & \textbf{28.08} \\
         Metric-depth & \cmark & \xmark & \underline{33.42} & \underline{32.63} &  \underline{28.14}\\
         Metric-depth & \cmark & \cmark & 37.18  &  36.28 & 30.18 \\
    \end{tabular}
    }%
    \caption{
    \textbf{FID with scale learning and trimming.}
    The average FID over a trajectory (21 frames, generated using keyframes \cite{polyoculus}); the last five, and last three frames to further show the effect of error accumulation.
    While trimming and scale learning have similar effects on SFC, trimming hurts FID.
    }
    \label{tab:keyframe-FID}
\end{table}

\section{Conclusion and Discussion}
In this paper we address the challenge of scale ambiguity in GNVS models inherent in multi-view datasets. 
We first demonstrated the problem, showing that scale ambiguities present in multiview training datasets manifest in the conditional image distributions learned by GNVS models.
Based on this we defined two metrics based on optical flow and epipolar geometry to quantify the scale ambiguity present in a GNVS model.
We further proposed a new method to learned scene scales by optimizing them during GNVS model training.
We evaluated this method and compared it to others, showing that the proposed method is effective at reducing the scale ambiguities in a trained GNVS model.
Further, we showed that image quality metrics were improved.
Our approach offers a simple yet effective and general solution that requires no preprocessing, and our metrics provide a reliable means of evaluating scale consistency in GNVS models.

\section{Acknowledgments}
This work was completed with support from the Vector Institute and was funded
in part by the Canada First Research Excellence Fund (CFREF) for the Vision: Science to Applications (VISTA) program (M.A.B., K.G.D., T.T.A.A.), the
NSERC Discovery Grant program (M.A.B., K.G.D.), and the NSERC Canada
Graduate Scholarship Doctoral program (J.J.Y.).
\clearpage
\newpage
{
    \small
    \bibliographystyle{ieeenat_fullname}
    \bibliography{refs_short, main}
}

\maketitlesupplementary
\setcounter{page}{1}
\setcounter{section}{0}
\setcounter{figure}{0}

\begin{abstract}
    This document provides additional material that is supplemental to the main submission.
    Section\ \ref{sec:sup-1} shows the effect of number of conditioning views on scale uncertainty.
    Section\ \ref{sec:sup-2} explains the mask generation process and
    Section\ \ref{sec:sup-3} discusses the need for optical flow normalization for SFC.
    Section\ \ref{sec:sup-4} discusses and quantifies the convergence of learned scales during training.
    Section\ \ref{sec:sup-5} details the specific method we use to estimate scene scales from monocular metric-depth.
    Section\ \ref{sec:sup-6} shows additional qualitative samples showing the effect of scale uncertainty on generated views.
    Section\ \ref{sec:sup-7} presents additional evaluations of scale variations and the effects of trimming.
\end{abstract}

\section{Scale Uncertainty as a Function of Conditioning Views}
\label{sec:sup-1}
\begin{figure}[t]
  \centering
   \includegraphics[width=0.75\linewidth]{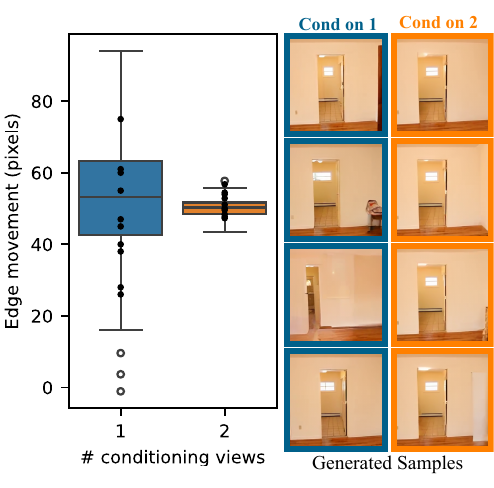}
   \caption{\textbf{Effect of the number of conditioning views on sample scale variance.} We take the right edge of the door and manually measure its movement in ten samples generated conditioning on one and two frames. We depict edge movements in both cases as boxplots. On the right, you can see four samples generated conditioned on one and two views, respectively.} 
   \label{fig: comparison}
\end{figure}

When generating a novel view using a single observed view from a model trained with uncalibrated data, a significant amount of scale uncertainty should be observed since the scale is underconstrained.
If the variation observed in the generation is primarily caused by scale uncertainty, then this additional entropy should be significantly reduced by introducing a second conditioning view to constrain the scale.
We perform a simple experiment where we generate novel views on a scene using PolyOculus~\cite{polyoculus} with pretrained weights, with one and two conditioning views.
As observed in Fig. \ref{fig: comparison}, the movement of a manually selected edge exhibits less variance across samples when using two conditioning views.
These results suggest that the variance in the novel views conditioned on single images is dominated by scale uncertainty and can potentially be measured by analysis of the generations.

\section{Details of SFC Mask Generation}
\label{sec:sup-2}
Some regions of the generated images may contain newly generated content with no correspondences in the conditioning view, and are invalid for measuring motion.
We compute occlusion masks to identify these regions using the well-known notion of \textit{forward-backward flow consistency}~\cite{ChangCC21, BlackA96}.
For a pair of frames A and B, consider a pixel $P_A$, in frame A. Then, the predicted forward flow, $f_{A \to B}$, should be consistent with the backward flow, $f_{B \to A}$, at the corresponding location at frame $B$ (at position $P_{A \to B} = P_A + f_{A \to B}$).
Hence, using $f_{A \to B}$ and $f_{B \to A}$, at each pixel location $P_A$ in frame $A$, we compute the absolute difference between the forward and backward flows at the corresponding location in frame $A$, defined as $d(P_A)$:
\begin{align}
\label{eq:for_back_diff}
    d(P_A) = ||f_{A \to B} (P_A) - f_{B \to A} (P_A + f_{A \to B})||_1.
\end{align}
The occlusion mask, $M$, is then generated by applying a threshold, $t$, on $d(P_A)$ at each pixel location, $(i,j)$: %
\begin{align}
    \label{eq:mask_cod}
    {M}_{ij} = 
        \begin{cases}
            0, & \begin{aligned}[t]
                &\text{if } ||d(p_A)||_\infty > t \\
                &\text{or } (P_A + f_{A \to B}) \text{ is out of bounds}
            \end{aligned} \\
            1, & \text{otherwise}
        \end{cases},
\end{align}

\section{Normalization in SFC}
\label{sec:sup-3}
\begin{figure*}[t]
   \includegraphics[width=\linewidth]{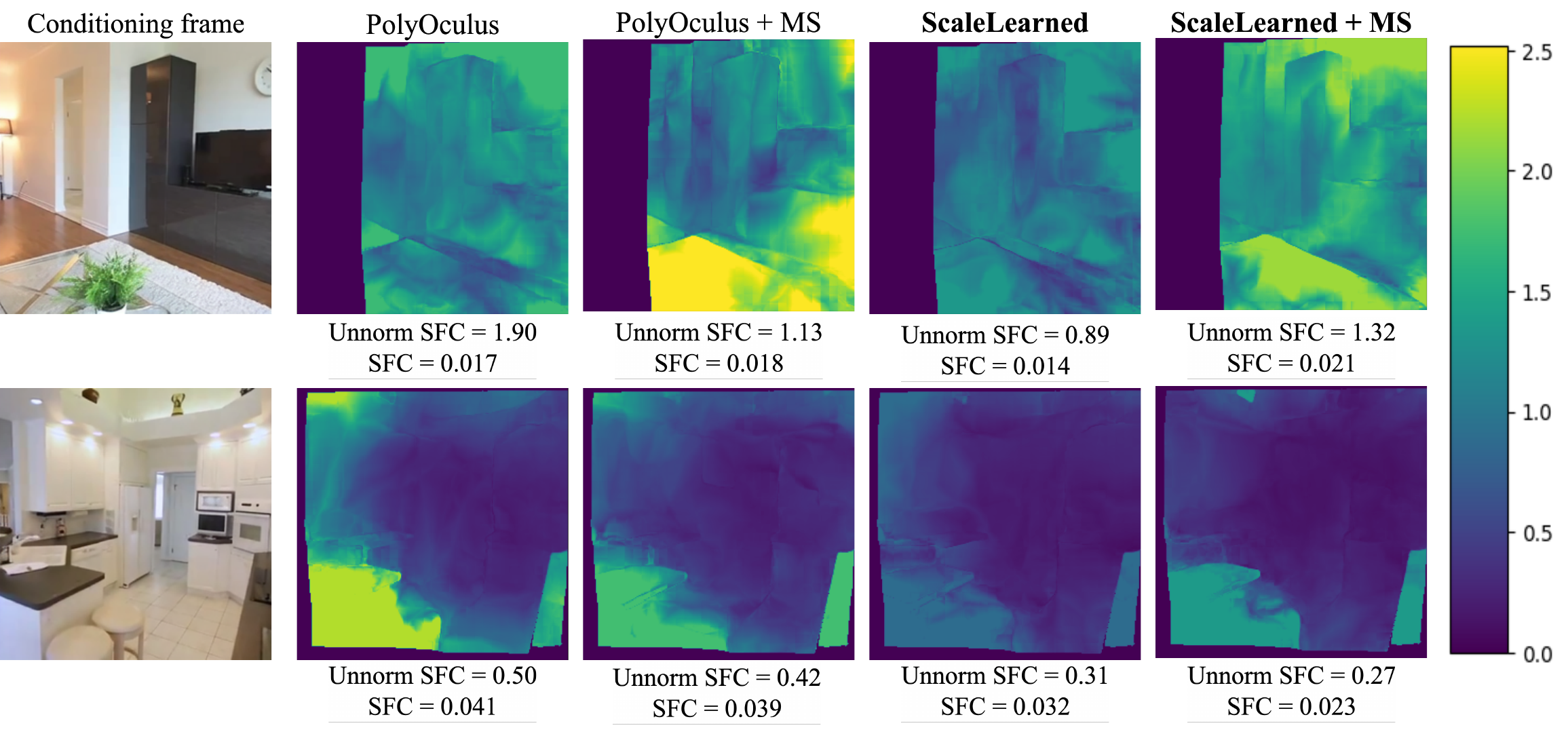}
   \caption{
   \textbf{Unnormalized per-pixel MAD flow maps of two scenes with significantly different motion magnitudes.}
   Unnormalized SFCs were computed with unnormalized MAD maps.
   (Top) Objects throughout this scene are generally closer to the camera, and the camera moves to the right.
   The combination of these factors results in motion with large magnitudes and variance in image space, yielding relatively light values in the MAD map visualization.
   (Bottom) Objects in this scene are generally further, and the the camera moves forward.
   This results in lower flow magnitudes and variance, yielding darker MAD maps.
   }
   \label{fig:sfc_norm}
\end{figure*}

As discussed in the main paper, the magnitude of motion optical flow varies substantially depending both on scene geometry, and camera motion.
In Figure\ \ref{fig:sfc_norm}, we visualize the unnormalized MAD maps of two scenes with significantly different flow magnitudes.
The variance of flow magnitudes between scenes necessitates normalization as part of SFC.

\section{Convergence of Learned Scales}
\label{sec:sup-4}
\begin{figure}
  \centering
   \includegraphics[width=\linewidth]{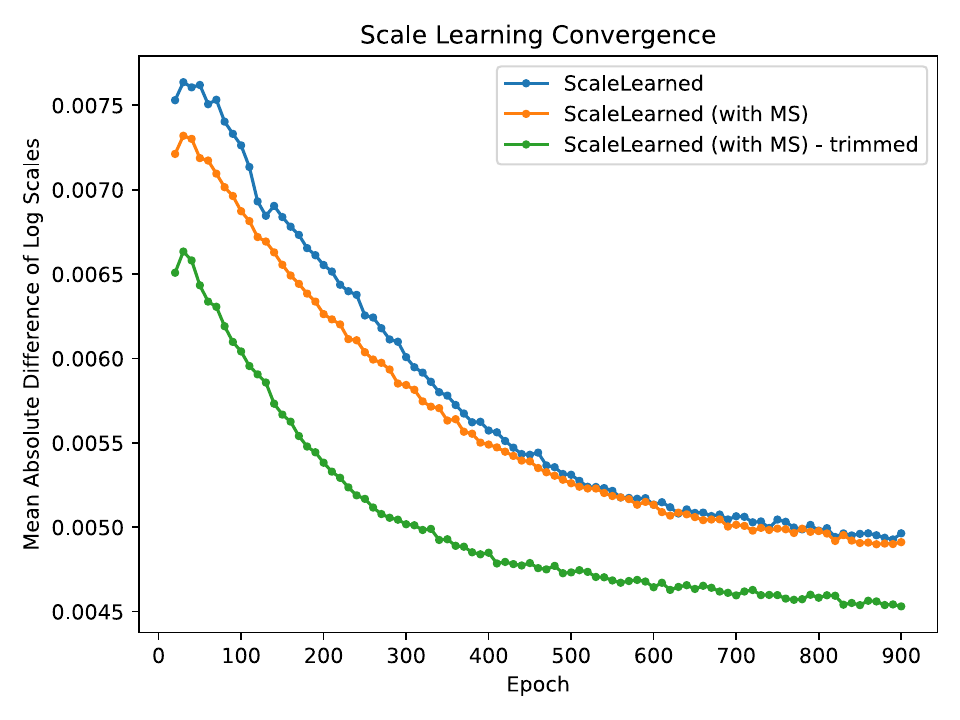}
   \caption{
   \textbf{Mean absolute difference of log scales plot.} This plot shows the trend of scale changes during training, and its plateau indicates scale convergence.}
   \label{fig:scale-convergence}
\end{figure}

\noindent\textbf{Rate of change during training.}\ 
To obtain a sense of convergence on the learned scales when the ground-truth scales are not available, we plotted average of absolute difference of log scales between every 10 epochs for models incorporating scale learning during training:
\begin{align}
    \Delta \text{scales}_k = \underset{i\in N}{\mathrm{Mean}}(|\log (\text{scales}_{i,k}) - \log(\text{scales}_{i,k-10})|),
\end{align}
where $k$ is the epoch, and this difference is averaged over $N$ scenes.
As shown in Figure \ref{fig:scale-convergence}, the differences plateau over time, which indicates the learned scales converging.

\begin{figure}
  \centering
   \includegraphics[width=\linewidth]{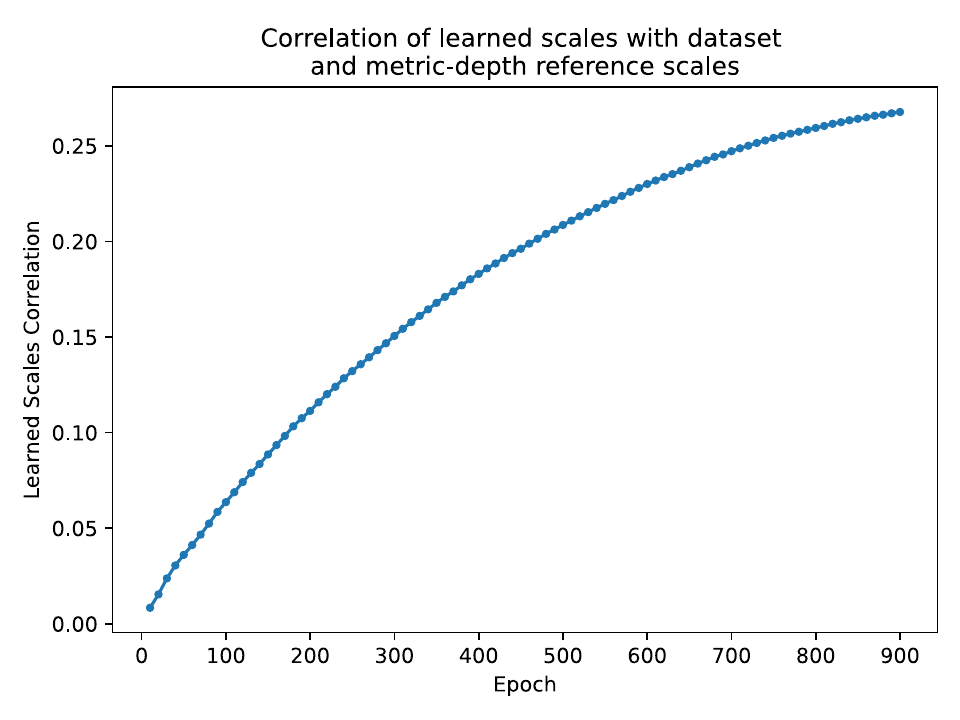}
   \vspace{-10pt}
   \caption{
   \textbf{Learned scales correlation.} Correlation curve between scales learned by the ScaleLearned models with dataset reference scales and scales learned by the ScaleLearned models with metric depth estimated reference scales with respect to training epoch.
   }
   \label{fig:scale-corr}
\end{figure}

\noindent\textbf{Convergence between methods with different scale initializations.}\ 
We also analyze the learned scales across models initialized with different reference scales.
Ideally, as each scene has a unique scale, models should independently converge to the same scale for a scene.
However, the scale learning process is noisy due to solely depending on the diffusion loss, and the optimization likely contains some local minima, resulting in different scales found by each model.
Figure\ \ref{fig:scale-corr} shows the correlation between the learned scales by the ScaleLearned model and ScaleLearned (with MS) increases during training, suggesting that both models are converging to similar scales with some independence from the reference scales.

\begin{table}
    \centering
    \resizebox{\linewidth}{!}{
    \footnotesize
    \begin{tabular}{ccc|ccc}
        \multicolumn{3}{c}{} & \multicolumn{3}{c}{FID $\downarrow$} \\
        \makecell{Reference \\scale} & \makecell{Scale \\learning} & Trim & All & Last 5 & Last 3 \\
        \hline
        Dataset & \xmark & -- & 39.40 & 57.16 & 59.70 \\
        Metric-depth & \xmark & \xmark & 42.01 & 62.03 & 65.08\\
        Metric-depth & \xmark & \cmark & 39.58 & 57.32 & 60.08\\
        Dataset      & \cmark & -- & \textbf{37.94} & \textbf{53.76} & \textbf{55.96} \\
        Metric-depth & \cmark & \xmark & \underline{38.15} & \underline{54.58} & \underline{56.94} \\
        Metric-depth & \cmark & \cmark &  39.62 & 56.75 &  58.97\\
    \end{tabular}
    }%
    \caption{
    Reporting FID on generated samples using Markov sampling spec for long generations.
    }
    \label{tab:markov-FID}
\end{table}

\section{Estimating Scene Scales Using Metric-Depth}
\label{sec:sup-5}
To estimate metric-depth reference scales, we followed the same procedure described in 4DiM~\cite{4DiM} to calibrate RE10K~\cite{realestate10k}.
We estimate dense monocular depth~\cite{depthanythingv2} from individual frames from the ground-truth videos.
We also recover sparse 3D points from the videos using COLMAP.
The metric-depth scene scale is estimated by least-square minimization of the depth of the sparse points estimated by COLMAP and monocular depth estimates for each view.
We also sort the scenes by the variance between the COLMAP and monocular depth estimates and consider scenes with the highest $~30\%$ variance as unreliable.
When training using all scenes, we set the scale of these unreliable scenes to $1.0$.

\section{Additional Examples of Scale Variability}
\label{sec:sup-6}
\begin{figure*}[t]
    \centering
    \includegraphics[width=0.9\linewidth]{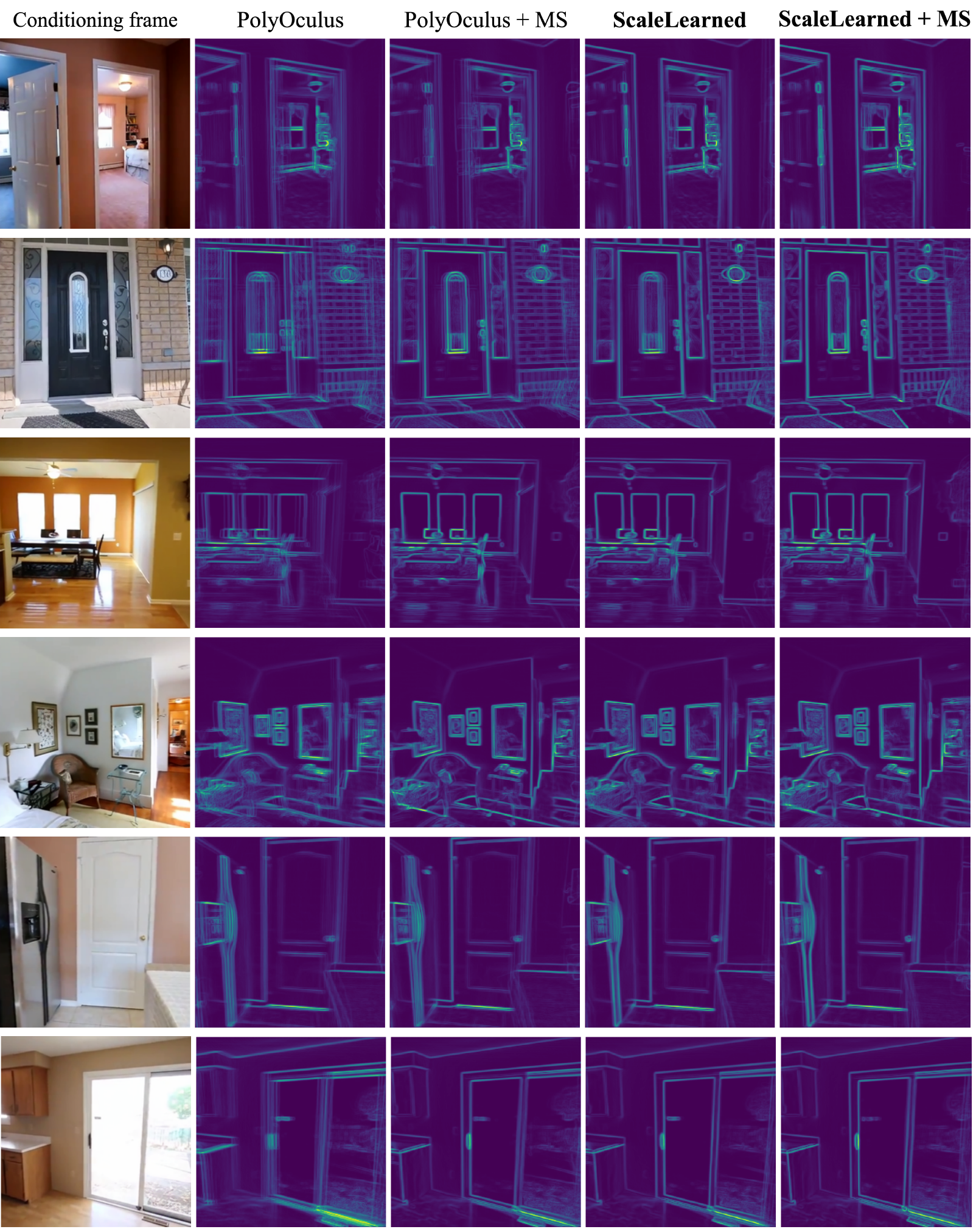}
    \caption{\textbf{Additional edge map visualizations.} On the left column, we depict the conditioning frame, and the heatmaps are generated by overlaying Sobel edge maps of samples generated with the same conditioning information.}
   \label{fig:edge_vis_suppl}
\end{figure*}

Here we provide additional qualitatives samples of overlayed edge-map visualizations under a fixed lateral motion, in Figure\ \ref{fig:edge_vis_suppl}.
Despite using identical camera motions, the displacement of the distinctive edge exhibits high variance due to the model's uncertainty about the scene scale.
In all cases, we can see edge locations stabilize with scale learning. Please refer to the \textbf{supplemental webpage} for an interactive viewer of these scenes where individual samples can be viewed.

\section{Additional Evaluations}
\label{sec:sup-7}
\subsection{Scale Variability Measurements with SFC}

\textbf{One direction translation trajectories.}
To assess the effect of scale learning in a more controlled setting, we also explore camera motions in which we only translate the camera along one of the axes without any rotation.
For this experiment, we use the same amounts of translation magnitude as explained in Sec. \ref{sec:experiments}.

\begin{figure*}[!t]
    \centering
    \begin{subfigure}[b]{0.33\textwidth}
         \centering
         \includegraphics[width=\linewidth]{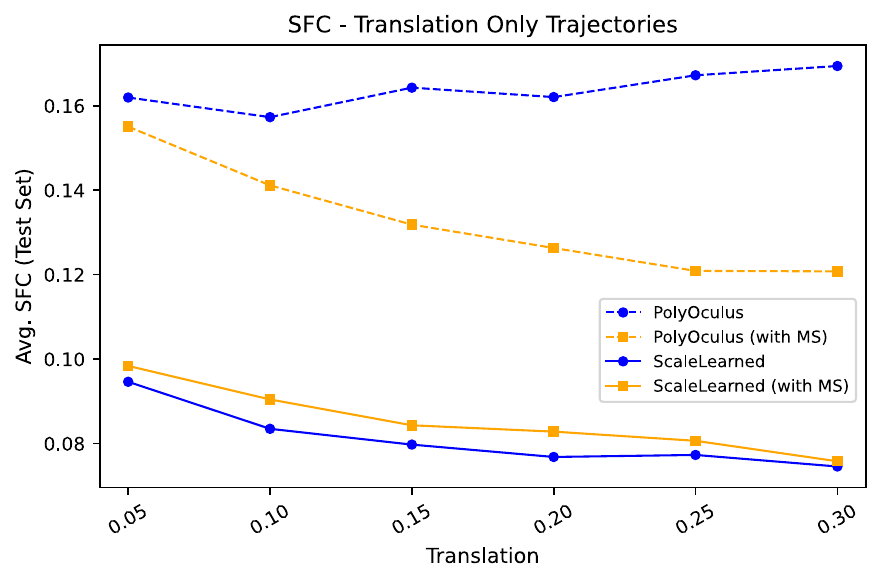}
    \end{subfigure}
    \begin{subfigure}[b]{0.33\textwidth}
         \centering
         \includegraphics[width=\linewidth]{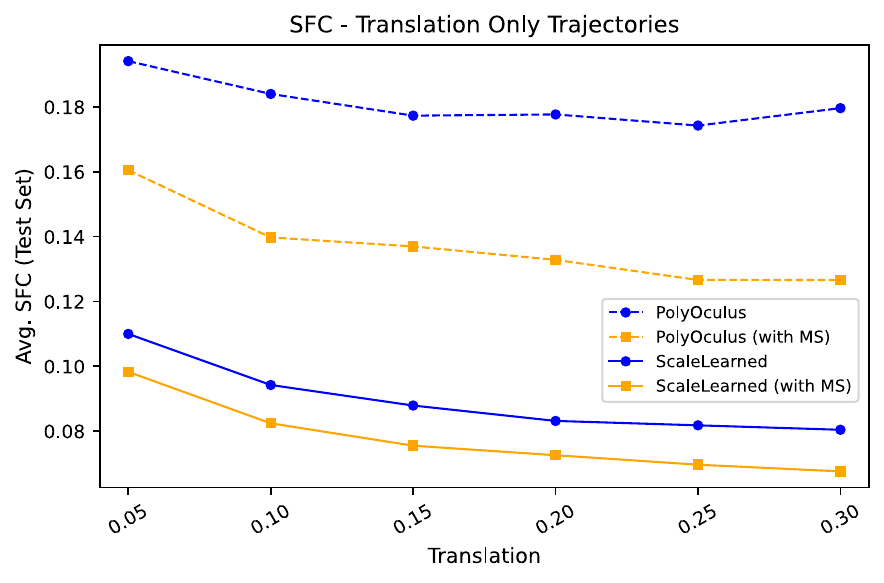}
     \end{subfigure}
     \begin{subfigure}[b]{0.33\textwidth}
         \centering
         \includegraphics[width=\linewidth]{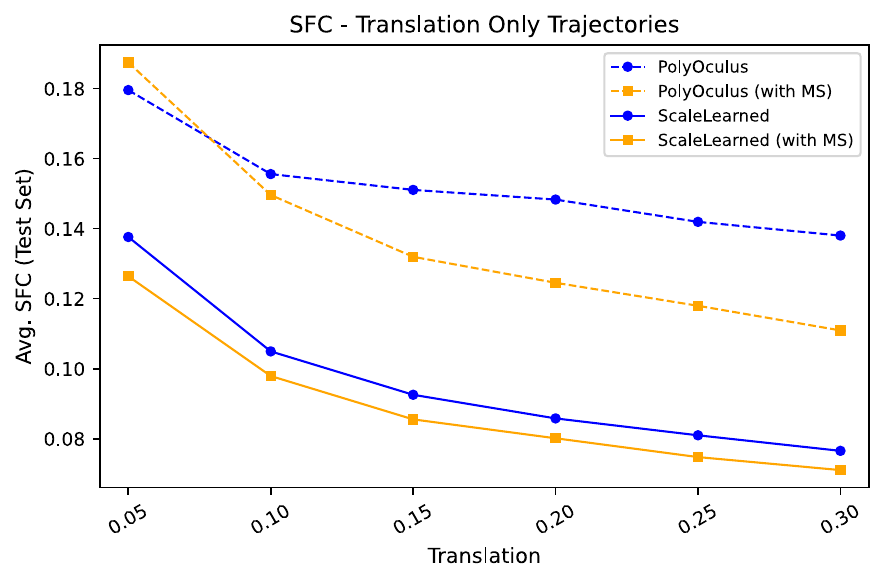}
     \end{subfigure}
    \caption{SFC results for samples with the camera motion only in one axis.}
    \label{fig:sfc_trans_only_each}

    \vspace{1em}  %

    \centering
    \begin{subfigure}[b]{0.33\textwidth}
         \centering
         \includegraphics[width=\linewidth]{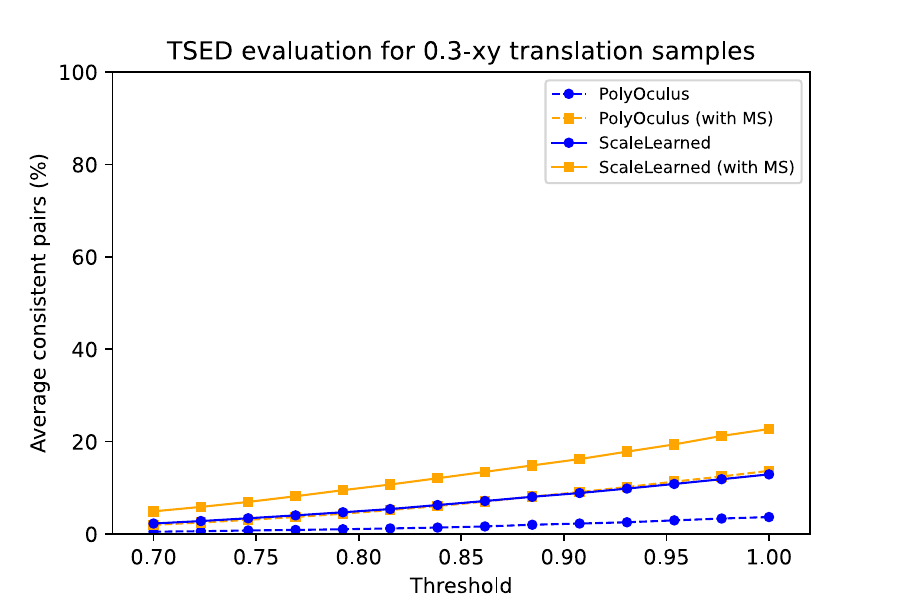}
    \end{subfigure}
    \begin{subfigure}[b]{0.33\textwidth}
         \centering
         \includegraphics[width=\linewidth]{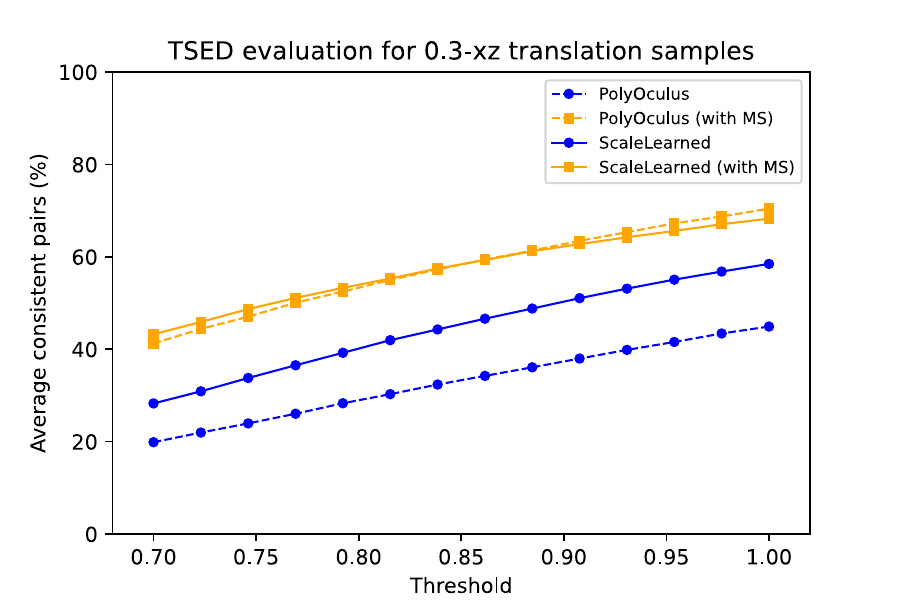}
     \end{subfigure}
     \begin{subfigure}[b]{0.33\textwidth}
         \centering
         \includegraphics[width=\linewidth]{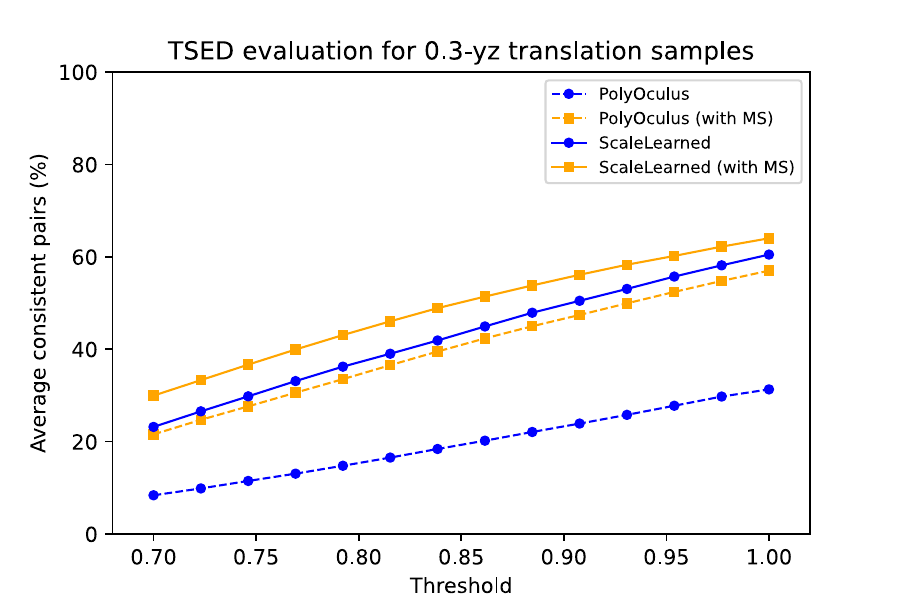}
     \end{subfigure}
    \caption{TSED results in distinct plots.}
    \label{fig:tsed_three_plot}
\end{figure*}

Due to the lack of ground truth frames in this setting, we generate SFC masks by comparing optical flows between the conditioning frame and the generated samples. Then, we aggregate all masks by averaging them. To ensure a fair comparison among models, we average masks generated by all models' samples so that all the models use the same mask in SFC computations. 
Figure \ref{fig:sfc_trans_only_each} shows that while metric-depth estimated scales help reduce sample scale entropy, incorporating scale learning further improves sample scale consistency. 
These results demonstrate that scale learning helps the model acquire a consistent scale in each 3D direction, and it is not limited to trajectories with the same distribution as the training set. 

\textbf{TSED experiment.} In figure \ref{fig:tsed_three_plot}, we show separate plots for each combination of axes (xy, xz, yz).  Results demonstrate that incorporating scale learning improves TSED across all reference scales.

\subsection{Effect of Trimming Unreliable Scenes}
We replicate the experiment from Section \ref{sec:results-trimming} using samples generated through Markov sampling rather than keyframing~\cite{polyoculus} and present the results in Table \ref{tab:markov-FID}.
These results highlight the increasing impact of error accumulation as sampling depth increases in Markov sampling compared to Keyframed setup.
Generally, the same trends in performance are observed here as with the experiments using Keyframed sampling.

\end{document}